\documentclass[11pt]{article}

\usepackage{UF_FRED_paper_style}
\usepackage{subcaption}
\usepackage{amsmath,amssymb,amsfonts}
\usepackage{algorithmic}
\usepackage{caption}
\usepackage{lscape}
\usepackage{graphicx}
\usepackage{pifont}
\usepackage{textcomp}
\usepackage{subcaption}

\usepackage{multirow}
\usepackage{xcolor}
\usepackage{float}
\usepackage{balance}

\usepackage{tabularx,booktabs}
\newcolumntype{C}{>{\centering\arraybackslash}X} 
\def\BibTeX{{\rm B\kern-.05em{\sc i\kern-.025em b}\kern-.08em
		T\kern-.1667em\lower.7ex\hbox{E}\kern-.125emX}}

\usepackage{lipsum}  %% Package to create dummy text (comment or erase before start)

\title{VQA with Cascade of \\Self- and Co-Attention Blocks\\}

\author{Aakansha Mishra\\% Name author
    \href{ak.kkb@iitg.ac.in}{\texttt{ak.kkb@iitg.ac.in}} %% Email author 1 
\and Ashish Anand\\% Name author
    \href{anand.ashish@iitg.ac.in}{\texttt{anand.ashish@iitg.ac.in}} %% Email author 2
\and Prithwijit Guha\\% Name author
    \href{pguha@iitg.ac.in  }{\texttt{pguha@iitg.ac.in}}%% Email author 3
    }
    
\date{}

\begin{document}
% %%%%%%%%%%%%%%%%%%%%%%%%%%%%%%%%%%%%%%%%%%%%%%%%%%%%%%%%%%
% %%%%%%%%%%%%%%%%%%%%%%%%%%%%%%%%%%%%%%%%%%%%%%%%%%%%%%%%%%
% ABSTRACT
% %%%%%%%%%%%%%%%%%%%%%%%%%%%%%%%%%%%%%%%%%%%%%%%%%%%%%%%%%%
% %%%%%%%%%%%%%%%%%%%%%%%%%%%%%%%%%%%%%%%%%%%%%%%%%%%%%%%%%%
{\setstretch{.8}
\maketitle
% %%%%%%%%%%%%%%%%%%
\begin{abstract}
The use of complex attention modules has improved the performance of the Visual Question Answering (VQA) task. This work aims to learn an improved multi-modal representation through dense interaction of visual and textual modalities. The proposed model has an attention block containing both self-attention and co-attention on image and text. The self-attention modules provide the contextual information of objects (for an image) and words (for a question) that are crucial for inferring an answer. On the other hand, co-attention aids the interaction of image and text. Further, fine-grained information is obtained from two modalities by using a Cascade of Self- and Co-Attention blocks (CSCA). This proposal is benchmarked on the widely used VQA2.0 and TDIUC datasets. The efficacy of key components of the model and cascading of attention modules are demonstrated by experiments involving ablation analysis.

\end{abstract}
}

% %%%%%%%%%%%%%%%%%%%%%%%%%%%%%%%%%%%%%%%%%%%%%%%%%%%%%%%%%%
% %%%%%%%%%%%%%%%%%%%%%%%%%%%%%%%%%%%%%%%%%%%%%%%%%%%%%%%%%%
% BODY OF THE DOCUMENT
% %%%%%%%%%%%%%%%%%%%%%%%%%%%%%%%%%%%%%%%%%%%%%%%%%%%%%%%%%%
% %%%%%%%%%%%%%%%%%%%%%%%%%%%%%%%%%%%%%%%%%%%%%%%%%%%%%%%%%%

% --------------------
\section{Introduction}
% --------------------
Initial attention-based approaches \cite{xu2016ask}\cite{kazemi2017show}\cite{yang2016stacked}\cite{anderson2018bottom} focused on identifying salient image regions based on the text of a given question. In other words, the focus was on giving attention to images (visual attention) only. Subsequent methods, often referred to as \textit{co-attention}-based methods \cite{lu2016hierarchical}\cite{ben2017mutan}, combined textual attention along with image attention. Textual attention focuses on relevant words in the context of the given image. Co-attention-based methods improved the performance of VQA systems. A few studies \cite{yang2016stacked}\cite{mishra2021multi}\cite{gao2019dynamic}\cite{gao2019multi} have shown that considering attention in a cascaded or stack-based manner helps in obtaining enriched representation with fine-grained information.

\begin{figure}[!t]
    \centering
    {\includegraphics[width=0.4\textwidth,height=8cm,keepaspectratio]{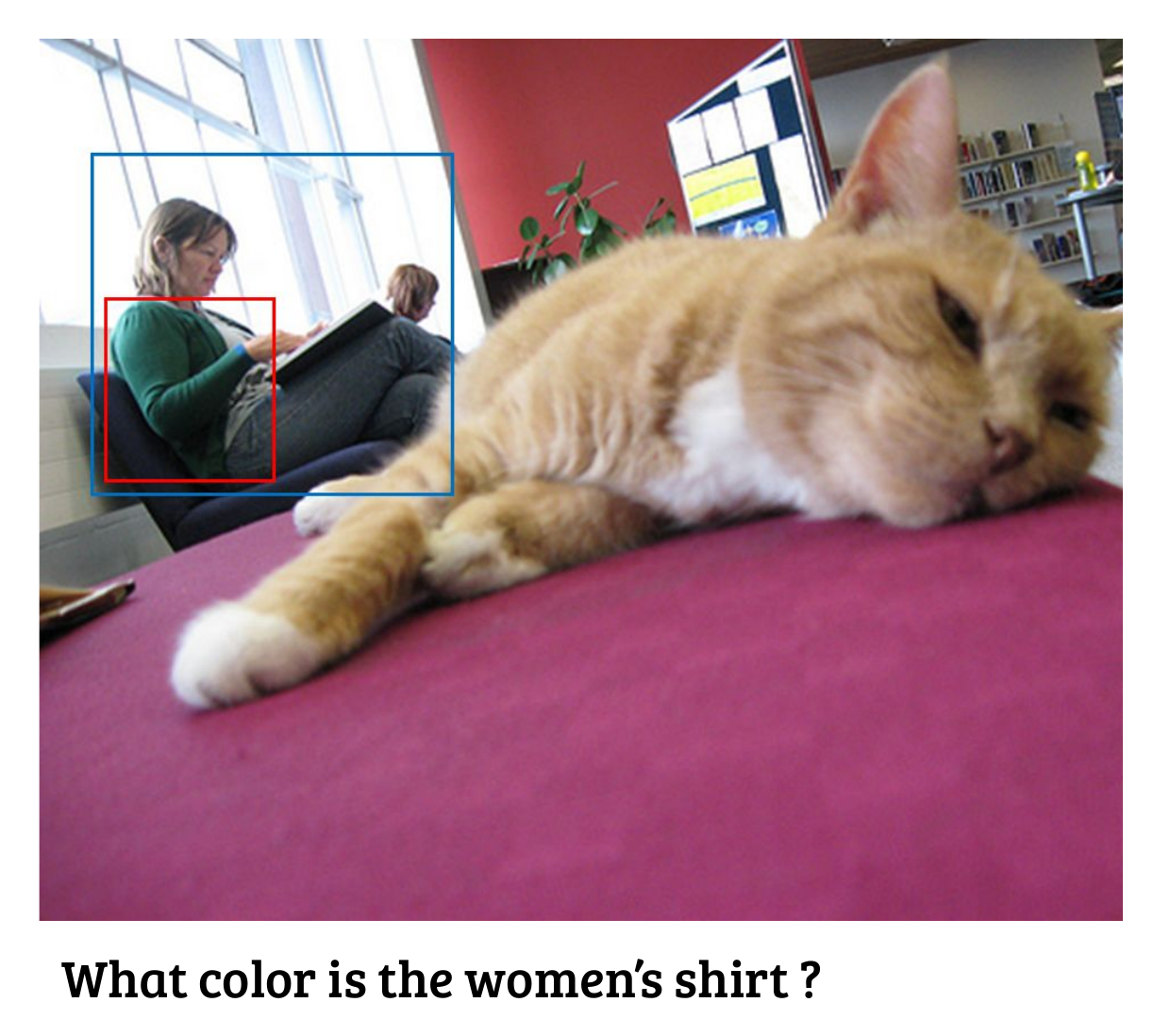}}
    \caption{{An example to illustrate the self-attention relevance for visual content.}}%
    \label{fig:overI}%
\end{figure}

Recent attention-based models have taken inspiration from transformer-based models \cite{vaswani2017attention} to include self-attention (SA) as well. The SA helps in incorporation of internal correlation within a modality. For text modality, SA encodes internal correlation among words to obtain informative representation of the given sentence. Similarly, for image modality, SA helps in encoding correlation among the salient regions of image. Figure \ref{fig:overI} shows an example for illustration. The given question is \textit{``What color is the women's shirt?}. Salient regions within image include `woman'. It is likely to be informative if the region consisting of `women' keeps the contextual information such as ``dress she is wearing", ``hair color" as well as correlation with other salient objects. Here, women's shirt could be one of the more correlated region with respect to some other salient objects. The SA helps in encoding such information.

\begin{figure}[!t]
    \centering
    {\includegraphics[width=0.4\textwidth,height=8cm,keepaspectratio]{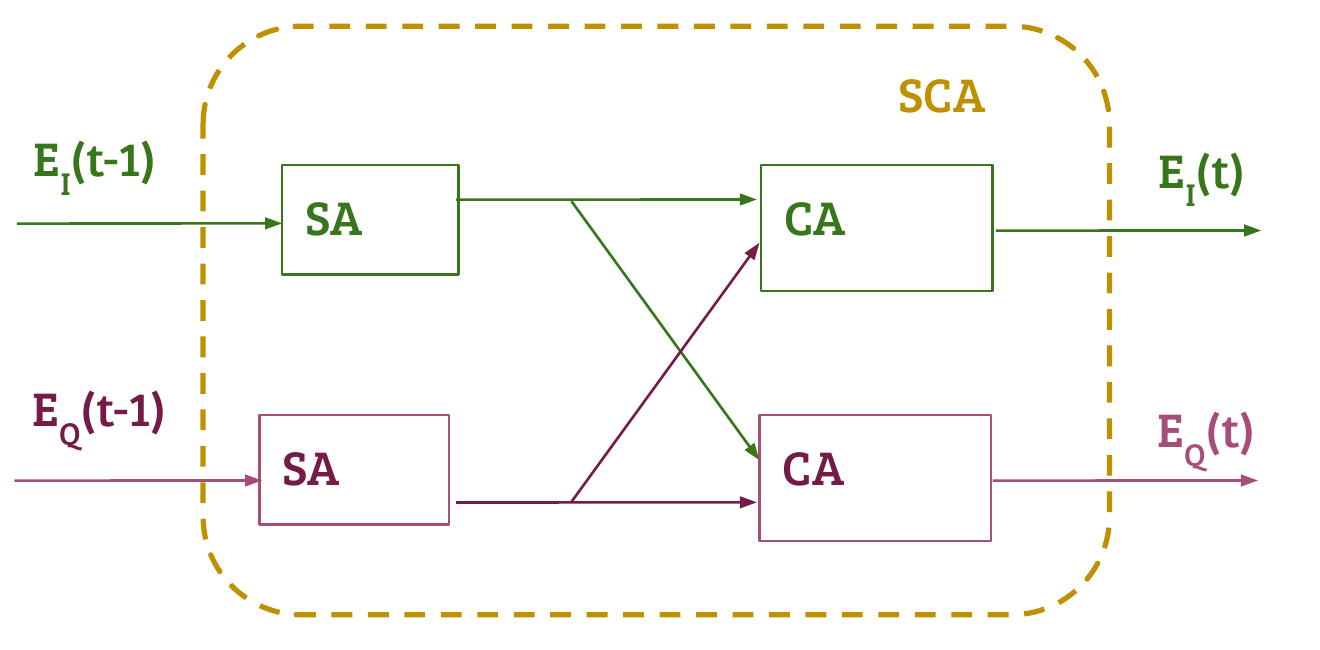}}
    \caption{{Overview of proposed model. An attention block, referred to as SCA, comprises of \textit{self-attention (SA)} and \textit{co-attention (CA)} modules. Multiple such attention blocks are cascaded, where output of some $(t-1)^{th}$ block is presented as input to the $(t)^{th}$ block.}}%
    \label{fig:overV}%
\end{figure}

Based on the advantages of each of the following modules: SA, co-attention (CA) and cascade of attention mechanisms, this work proposes combining them together in a systematic manner. Towards this objective, the proposed model builds one self- and co-attention based attention block (SCA), that combines both SA and CA in a specific way. For each of the text and image modalities, a specific SA module obtains a feature representation for the respective modality. Then the co-attention module uses self-attended representation of one modality and attends (takes attention) on the self-attended representation of the other modality to obtain cross-modality contextual representation for the second modality. Thus, there are two SA modules (one for each text and image modalities) and two co-attention modules within a single SCA block (Figure~\ref{fig:overV}). In one complex attention block of SCA, both modalities guide itself to capture internal correlation and each other to learn the robust representation of each of the visual and textual domains.

The proposed model exploits the niche attributes of the different attention mechanisms and further combines them together in a dense attention module (SCA block). A Cascade of multiple SCA blocks (CSCA) is used to extract fine-grained information. Figure~\ref{fig:overV} gives the overview of the $t^{\text{th}}$ SCA block which takes representations of question and image of the $(t-1)^{\text{th}}$ block as input, and provides the improved representation of question and image.

To analyse and evaluate the model performance, extensive experiments are performed on two widely used VQA datasets: \textit{VQA2.0} \cite{goyal2017making} and \textit{TDIUC} \cite{kafle2017analysis}. Ablation analysis experiments are also performed to understand the impact of the important components of the proposed model. Primary contributions of this work are:

\begin{itemize}

	\item A dense attention based VQA model comprising of cascaded attention blocks. 
	
	\item The core of each attention block consists of self-attention and co-attention so that the two modalities guide each other to obtain an enriched representation.
	
	\item Extensive performance evaluation along with ablation analysis of the proposed model on the two benchmark datasets -- \textit{TDIUC} and \textit{VQA2.0}.

\end{itemize}

% Related Work
\section{Related Work}
\label{sec:relwork}
VQA, being a multimodal task, requires an unified representation of the text and image modalities. Initial VQA models \cite{antol2015vqa,goyal2017making,simonyan2014very,he2016deep} adopted simple fusion based approaches. These models first obtained feature representations of individual modality using corresponding pre-trained networks and then combined them to obtain a joint representation using a fusion schema. Simple fusion schemes include concatenation or element-wise summation or multiplication. Fukui et al. \cite{fukui-etal-2016-multimodal} proposed bi-linear pooling to capture interaction of components of the two modalities in a better way. Seeing the advantage of the bilinear pooling based fusion methods, further variants of bilinear pooling with lesser complexity or faster convergence were proposed. MFB~\cite{yu2017multi}, MLB \cite{kim2016hadamard}, MFH~\cite{yu2018beyond} were proposed to obtain a representation providing better interaction of the two modalities.

Introduction of attention mechanism in \cite{attention15} equipped neural models with a systematic procedure to assign relative weights of importance to sequential inputs. Shi et al. in \cite{shih2016look} have introduced image attention guided by question to focus on salient image regions relevant to the given question. This helped in obtaining improved feature representations. This led to the development of several attention based approaches for VQA \cite{kazemi2017show}\cite{kim2018bilinear}\cite{yang2016stacked}\cite{wu2018chain}\cite{mishra2021multi}\cite{xu2016ask}\cite{anderson2018bottom}. Studies in \cite{yang2016stacked}\cite{wu2018chain}\cite{mishra2021multi} have shown that applying attention multiple times helps in obtaining enriched representation embedded with fine-grained information. 

\begin{figure*}[!h]
\center
\includegraphics[width=1\textwidth,height=8cm,keepaspectratio]{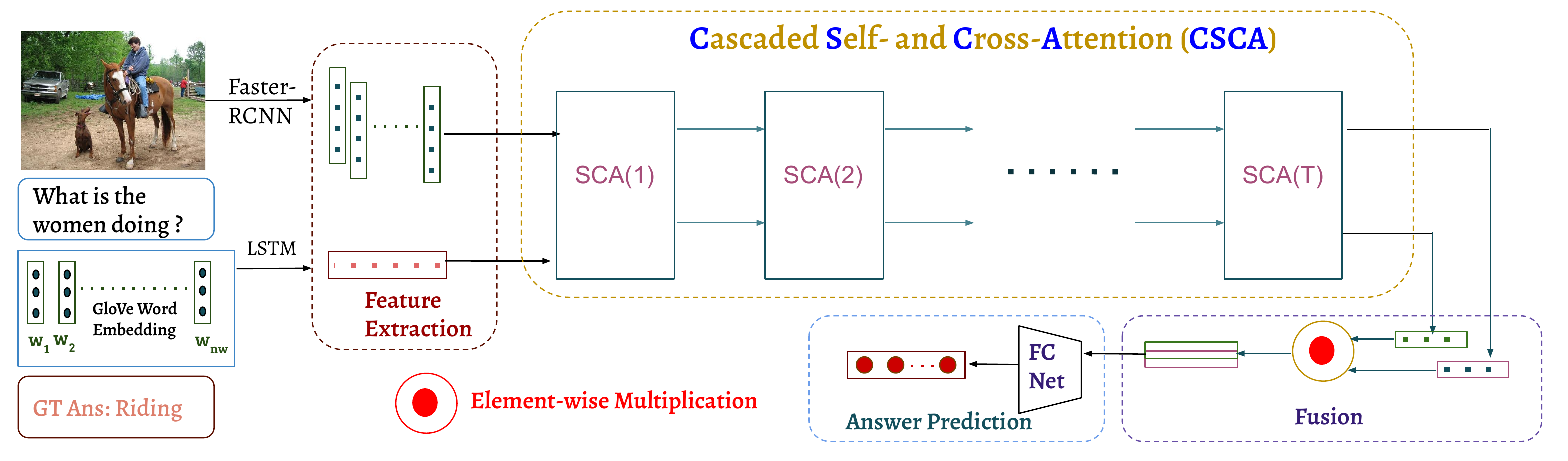}
\caption{Functional block diagram of the proposed approach. Initial feature extraction followed through a cascade of self-attention and co-attention mechanisms. Final features are fused through element-wise multiplication and fed to a fully connected network for answer classification.} 
\label{fig:framework}
\end{figure*}

Authors in \cite{lu2016hierarchical}\cite{yu2017multi} have proposed that attention on textual features in context of visual features along with visual attention plays a key role in VQA models. Such two way attention mechanism is referred to as \textit{dual attention} or \textit{co-attention} or \textit{cross-modality attention} in the literature. We have also used these terms interchangeably. Kim et al. \cite{kim2018bilinear} have proposed bilinear interaction based attention for dual modality. Do et al. \cite{do2019compact} have proposed an approach by exploiting knowledge distillation with a teacher and student model. Mishra et al. \cite{mishra2021multi} have proposed co-attention based multistage model for VQA. In another work, the authors \cite{mishra2022dual} have proposed question categorization and dual attention for VQA. RAMEN \cite{shrestha2019answer} is an unified model that uses high level reasoning and can deal with VQA datasets based on both real-world and synthetic images.

Another class of attention mechanism uses intra-modal attention (self-attention) along with cross-modal attention (co-attention) to learn better feature representation. Gao et.al. \cite{gao2019dynamic} have proposed DFAF that combines self-attention and co-attention. Multi-modal Latent Interaction (MLIN) \cite{gao2019multi} used multi-modal reasoning through summarization, interaction, and aggregation. Yu et al. \cite{yu2019deep} have proposed an encoder-decoder based dense attention mechanism. These models are relatively dense than the previous approaches and hence, are referred to as dense attention based models. Authors in \cite{lu2019vilbert}\cite{tan2019lxmert} have proposed transformer based attention models for multimodality tasks. These models are pretrained for multiple tasks on huge datasets, that could be further exploited for downstream tasks.

The proposed model falls in the category of dense attention based methods. It uses a cascade of attention blocks to obtain a multi-modal feature representation. Here, each attention block comprises of intra-modality and cross-modality interactions. The proposed method is described next.

% Proposed Method
\section{Proposed Method}
\label{sec:propMeth}

The proposed framework treats VQA as an answer classification task following existing works like \cite{anderson2018bottom}\cite{gao2019dynamic}\cite{antol2015vqa}\cite{goyal2017making}\cite{gao2019multi}. The input image $I$ ($I \in \mathbf{\mathcal{I}}$) and the associated natural language question $q$ ($q \in \mathbf{\mathcal{Q}}$) are first subjected to feature extraction (Subsection~\ref{subsec:FeatExtract}). Pretrained deep networks are used to extract features from a few salient image regions. The network embeddings are used to represent the input image. Similarly, a pretrained network is used to obtain the word embeddings of the associated input question. These word embeddings collectively represent the input question. The feature embeddings of both image and text modalities are subjected to self-attention mechanism (Subsection~\ref{subsec:saMod}) for capturing the relationships among different regions of $I$ and words of $q$. The self-attended representations of these two modalities are further processed by co-attention modules (Subsection~\ref{subsec:caMod}). This single stage of \textbf{S}elf and \textbf{C}o-\textbf{A}ttention mechanism cascade forms a single SCA block (Figure~\ref{fig:overV}). Multiple SCA blocks are cascaded to obtain further fine grained representations of both modalities. The embeddings obtained from the final SCA block are fused (Subsection~\ref{subsec:stacFus}) and fed to the answer classification network (Subsection~\ref{subsec:ansClass}) to predict the answer $\hat{a}$ ($\hat{a} \in \mathbf{\mathcal{A}}$).

% Feature Extraction
\subsection{Feature Extraction}
\label{subsec:FeatExtract}

% Visual Feature Extraction
A pretrained deep network based object detection model (Faster R-CNN, \cite{ren2015faster}) is used to identify the top-$n_v$ salient regions from the input image $I$. The pretrained ResNet-101 \cite{he2016deep} network is used to compute the visual feature of each region as an embedding $\mathbf{r} \in \mathbb{R}^{d_{v}}$. Thus, the input image $I$ is represented as $\mathbf{rI} \in \mathbb{R}^{d_v \times n_v}$ by using $n_v$ number of $d_v$ dimensional ResNet-101 embeddings.

\begin{equation}
   \mathbf{rI} = [ \mathbf{r}_1, \ldots \mathbf{r}_{n_v} ]; \mathbf{r} \in \mathbb{R}^{d_v}  
\end{equation}

% Question Feature Extraction
The input natural language question $q$ is first padded and trimmed to a length of $n_w$ words. The word features are further extracted as pretrained GloVe embeddings $\mathbf{eq} \in \mathbb{R}^{d_w}$ \cite{pennington2014glove}. Thus, the question $q$ is represented as $\mathbf{E_{q}} \in \mathbb{R}^{d_w \times n_w}$ by using $n_w$ number of $d_w$ dimensional embeddings.

\begin{equation}
   \mathbf{E_{q}} = [ \mathbf{eq}_1, \ldots \mathbf{eq}_{n_w} ]; \mathbf{eq} \in\mathbb{R}^{d_w} 
\end{equation}

All feature embeddings in $\mathbf{rI}$ and $\mathbf{E_{q}}$ are projected to a common $d$ dimensional space to obtain the respective initial feature embedding matrices as $\mathbf{rI}(0)$ and $\mathbf{E_{q}}(0)$.

\begin{eqnarray}
   \mathbf{\mathbf{rI}(0)} & = & W_c^{I}\mathbf{rI}\\
   \mathbf{\mathbf{Eq}(0)} & = & W_c^{Q}\mathbf{E_q}
\end{eqnarray}

\noindent Here, $W_c^{I} \in \mathbb{R}^{d \times d_{v}}$ and $W_c^{Q}\in \mathbb{R}^{d \times d_{w}}$ are the transformation matrices. These representations are provided as input to the self- and co-attention modules.

% Self Attention
\subsection{Self-Attention}
\label{subsec:saMod}

The self-attention (SA) mechanism is one of the key components of the proposed model. It is incorporated for both textual (question as collection of words) and visual (image as top-$n_v$ salient regions) modalities. At the $t^{\text{th}}$ ($t=1, \ldots T$) block, the input to SA are $\mathbf{rI}(t-1)$ and $\mathbf{E_{q}}(t-1)$. Following \cite{vaswani2017attention}, the SA uses \textit{keys} and \textit{queries}, both of dimension $d_{KQ}$ and values of dimension $d_{VS}$ respectively. The \textit{Multi-Head Attention} \cite{vaswani2017attention} is incorporated to capture the attention from different aspects. For this, $n_h$ parallel heads are added, where each head is considered to learn the relationships from different view (for image) and context (for question). 

Let $\mathbf{E_M} = \{ \mathbf{em}_1 \ldots \mathbf{em}_l \}$ be a matrix of feature embeddings, where $\mathbf{em} \in \mathbb{R}^ {d_m}$ and $\mathbf{E_M} \in \mathbb{R}^ {d_m \times l}$. For visual features, $\mathbf{E_M} = \mathbf{rI}(t-1)$, $l = n_v$ and $d_m = d$. Similarly, for question features, $\mathbf{E_M} = \mathbf{E_q}(t-1)$, $l = n_w$ and $d_m = d$. 

The query ($Q_S^{(i)}$), key ($K_S^{(i)}$) and value ($V_S^{(i)}$) matrices for the $i^{\text{th}}$ head can be respectively expressed as follows
\begin{eqnarray}
    Q_S^{(i)} & = & \left( W_i^{QS} \right)^\intercal \mathbf{E_M} \\
    K_S^{(i)} & = & \left( W_i^{KS}\right)^\intercal \mathbf{E_M} \\
    V_S^{(i)} & = & \left( W_i^{VS}\right)^\intercal \mathbf{E_M}
\end{eqnarray}

\noindent where, $W^{QS}_i \in \mathbb{R}^{d_m \times d_{KQ}}$, $W^{KS}_i \in \mathbb{R}^{d_m \times d_{KQ}}$ and $W^{VS}_i \in \mathbb{R}^{d_m \times d_{VS}}$ are transformation matrices. Using $\{ Q_S^{(i)}, K_S^{(i)}, V_S^{(i)} \}$, the inner product of query is performed with all the keys and is divided by $\sqrt[]{d_k}$ for more stable gradients \cite{vaswani2017attention}. The \emph{SoftMax} function is applied on the inner product to obtain the attention weights for question words and image salient regions. A scaled inner product based attention is computed for all the heads in the following manner.
\begin{figure}[!t]
\centering
\includegraphics[width=0.8\linewidth, height=0.4\textheight, keepaspectratio]{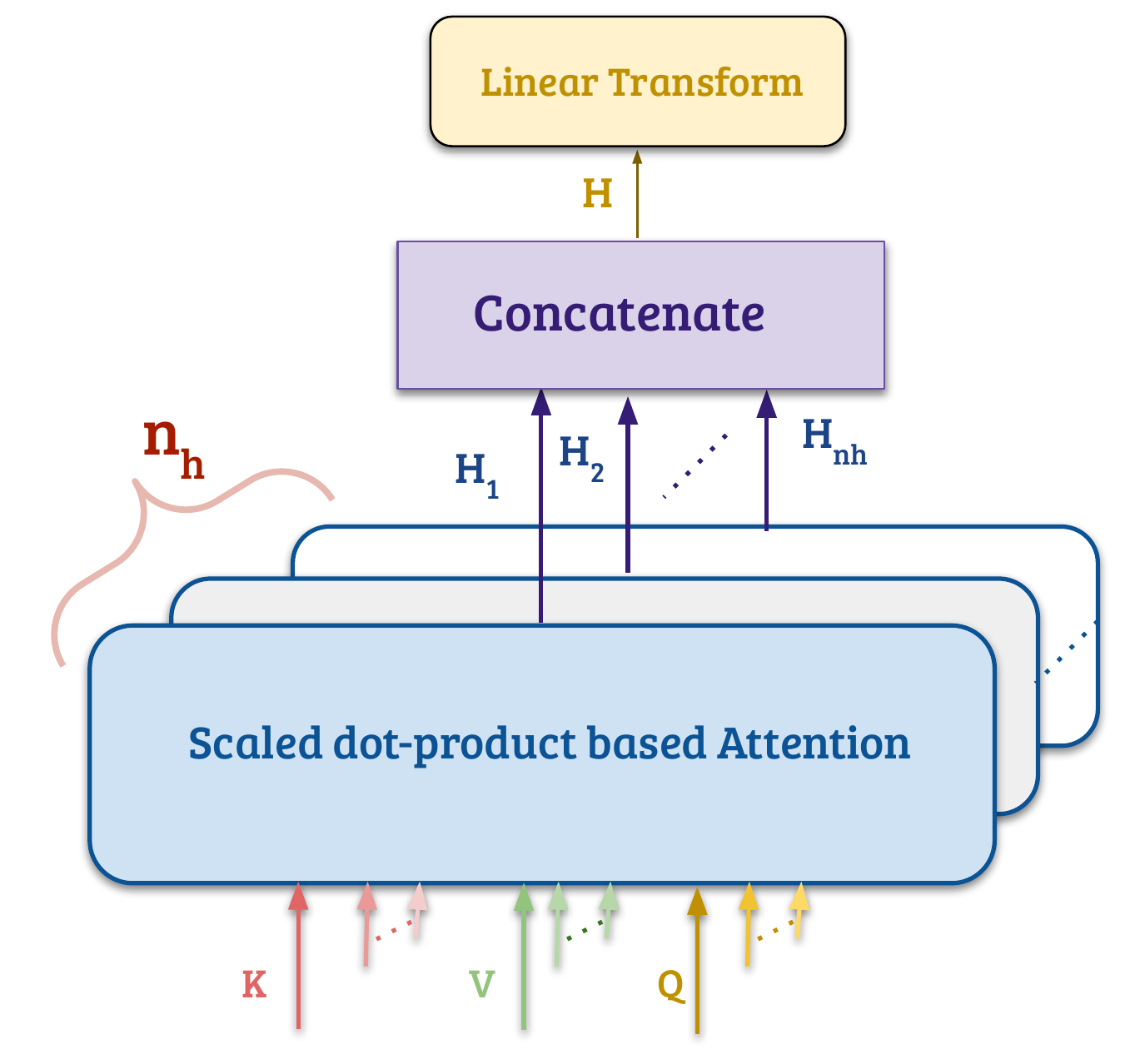}
\caption{Multihead Attention Mechanism}
\label{fig:mHead}
\end{figure}
\begin{equation}
     \mathbf{H_i} =
    \left(V_S^{\text(i)}\right)\mathrm{SoftMax}\left(\frac{{Q_S^{\text(i)}}^\top K_S^{\text(i)}}{\sqrt[]{d_K}} \right)
    \label{eq:scDot}
\end{equation}
\begin{equation}
    \mathbf{MH}( \mathbf{E_M} ) = W_{mh}\mathbf{H}
    \label{eq:mhead}
\end{equation}

\noindent Here, $W_{mh} \in \mathbb{R}^{ d_m \times (n_h \times d_{VS}) }$ is the transformation matrix. The output ( $\mathrm{MH}(\mathbf{E_M})$ ) of multi-head attention module is passed through fully connected feed forward layers with ReLU activation and dropout to prevent overfitting. Further, residual connections \cite{he2016deep} followed by layer normalization are applied on top of fully connected layers for faster and more accurate training. The layer normalization is applied over the embedding dimension only. Finally, the self-attended embeddings of the input feature $\mathbf{E_M}$ are obtained as $\mathbf{SE_M} = \{ \mathbf{sem}_{1} \ldots \mathbf{sem}_{l}\}$ where $\mathbf{sem} \in\mathbb{R}^{d_{m}}$ and $\mathbf{SE_M} \in \mathbb{R}^ {d_m \times l}$. Multihead attention mechamism is shown in Figure~\ref{fig:mHead}.
% Co-Attention
% \vspace{-0.3in}
\begin{figure}[!htbp]
% \vspace{-0.3in}
\centering
\includegraphics[width=\linewidth, height=0.4\textheight, keepaspectratio]{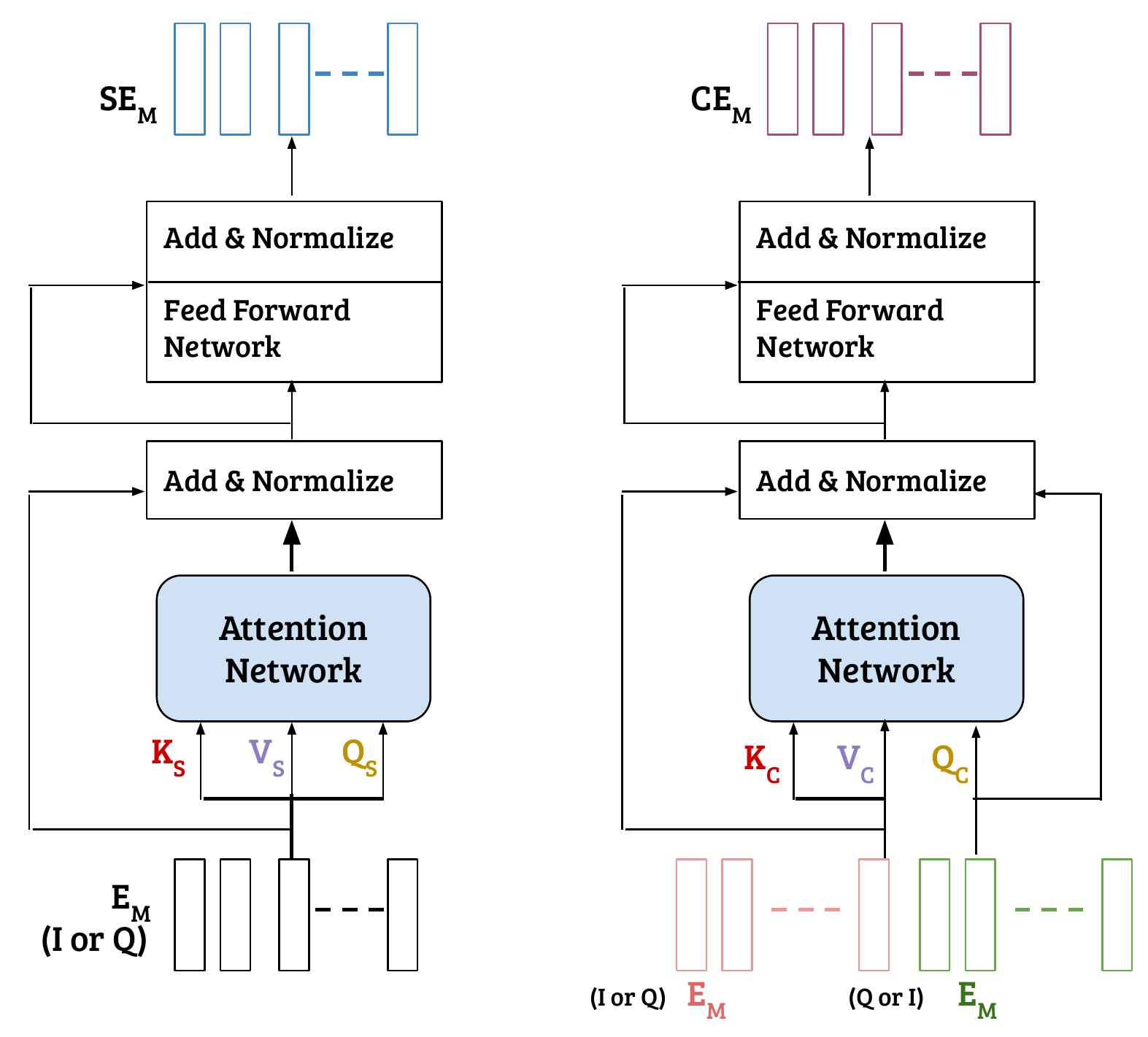}
\caption{\textit{Self-attention} and \textit{Co-attention} mechanism overview. Here, $M$ denotes the input modality.}
\label{fig:scAtt}
\end{figure}
\subsection{Co-Attention}
\label{subsec:caMod}
For cross-modal interactions, the co-attention module intakes the representations of two modalities and generates attention in context of each other. To facilitate this, the self-attended embeddings $\widetilde{\mathbf{E_q}}(t-1)$ and $\widetilde{\mathbf{rI}}(t-1)$ are taken as input. For generating image attention in context of question words, keys and values are generated from self-attended intermediate question representation while the query is obtained from the image itself (following Equation~\ref{eq:scDot}). Thus, the query ($Q_C^{(i)}$), key ($K_C^{(i)}$) and value ($V_C^{(i)}$) are respectively computed as follows.
\begin{eqnarray}
    Q_{C}^{(i)} & = & \left(W_i^{QC}\right)^\intercal \widetilde{\mathbf{E_q}}(t-1)\\
    K_{C}^{(i)} & = & \left(W_i^{KC}\right)^\intercal \widetilde{\mathbf{rI}}(t-1)\\ 
    V_{C}^{(i)} & = & \left(W_i^{VC}\right)^\intercal \widetilde{\mathbf{E_q}}(t-1)
\end{eqnarray}

\noindent Here, $W_i^{QC} \in \mathbb{R}^{d_m \times d_{KQ}}$, $W_i^{KC} \in \mathbb{R}^{d_m \times d_{KQ}}$ and $W_i^{VC} \in \mathbb{R}^{d_m \times d_{KV}}$ are transformation matrices. Similarly, for cross-modal question attention, the query is obtained from self-attended question embeddings. While the keys and values are obtained from self-attended image embeddings. These queries, keys and values are similarly processed following Equations~\ref{eq:scDot} and ~\ref{eq:mhead} to obtain the multi-head attention. This is fed to fully connected layers with ReLU, dropout, skip connections and layer normalization. The output of this network provides the final output of the co-attention module. Figure~\ref{fig:scAtt} demonstrates the overview of the \textit{self-attention} and \textit{co-attention} mechanism followed.

% Stacking and Fusion
\subsection{Cascading \& Fusion}
\label{subsec:stacFus}
A single SCA block comprising of \textit{self-attention} (intra-modality interaction) and \textit{co-attention} (inter-modality interaction) generates an enriched representation ($\mathbf{rI}(t),\mathbf{E_q}(t)$) of its input visual and textual features.

Existing works \cite{yang2016stacked}\cite{mishra2021multi} suggest the stacking of multiple such blocks to obtain further fine grained representations. This is accomplished by cascading multiple SCA block to $T$ steps. Let $\mathbf{rI}(T) \in \mathbb{R}^{d \times n_v}$ and $\mathbf{E_q}(T) \in \mathbb{R}^{d \times n_w}$ be the respective visual and question representations obtained from the final ($T^{\text{th}}$) SCA block.

The feature representations are obtained by averaging the attended embeddings of two modalities. So, the final visual embedding, say $\mathbf{I}_f$ is obtained as follows.
\begin{eqnarray}
  \mathbf{I}_{f} = \dfrac{1}{k}\sum_{j = 1}^{k} \mathbf{rI}(T)[:,j]  
\end{eqnarray}

\noindent Similarly, the question encoding, say $\mathbf{Q}_f$ is evaluated in the following manner.
\begin{eqnarray}
  \mathbf{Q}_{f} = \dfrac{1}{n_w}\sum_{j = 1}^{n_w} \mathbf{E_q}(T)[:,j]
\end{eqnarray}

The unified multi-modal representation, say $\mathbf{F} \in \mathbb{R}^d$ is obtained by fusing $\mathbf{I}_f$ and $\mathbf{Q}_f$ through element-wise multiplication.
\begin{eqnarray}
    \mathbf{F} = \mathbf{I}_f \odot \mathbf{Q}_f
    \label{eq:fusEmb}
\end{eqnarray}

The fused embedding $\mathbf{F}$ is fed to a fully connected network for answer prediction.

% Answer Classification and Model Learning
\subsection{Answer Prediction}
\label{subsec:ansClass}

The fused embedding $\mathbf{F}$ is fed to fully connected network with single hidden layer of dimension $d_{hp}$. The number of labels at the output layer is $n_c$ ($n_c = \mid \mathbf{\mathcal{A}} \mid$). The output answer vector, say $\mathbf{\hat{a}}$ is predicted as follows.
\begin{eqnarray}
    \mathbf{\hat{a}} = \mathrm{FCNet}\left( \mathbf{F} ; d_{hp}; n_c\right)
\end{eqnarray}
\subsection{Model Learning}
\label{sub:modLearn}

Let the respective ground truth and predicted answer be $a$ and $\hat{a}$ ($a,\hat{a} \in \mathbf{\mathcal{A}}$) for input image $I$ and question $Q$. This model uses cross-entropy loss for answer prediction and is defined as

\begin{equation}
    \mathcal{L}_c = -\sum_{j=1}^{n_c} a[j] log(\hat{a}[j])
\end{equation}

\noindent The combined set of parameters for proposed model includes the ones for feature extraction, block of dense attention and fusion mechanism. 

% Model Learning
\section{Experiment Design}
\label{sec:expt}
This section discusses the datasets used to benchmark the proposed model, the three evaluation metrics and the necessary implementation details.

\subsection{Dataset} 
\label{subsec:DS}
The proposed model is evaluated through experiments performed on the datasets VQA2.0 \cite{goyal2017making} and TDIUC \cite{kafle2017analysis}. The VQA2.0 \cite{goyal2017making} dataset is widely used for the VQA task. There are three question categories in VQA2.0. These are \textit{`Yes/No'} ($37.6\%$), \textit{`Number'} ($13.03\%$) and \textit{`Other'} ($49.37\%$). The dataset is divided into \textit{train}, \textit{validation} and \textit{test} sets with $443757$, $214354$ and $447793$ image, question and answer triplets respectively. 

The Task-Directed Image Understanding Challenge (TDIUC) \cite{kafle2017analysis} is another large VQA dataset of real images. Questions are categorized into $12$ types. These are \textit{`Scene Recognition'} ($4.03\%$), \textit{`Sport Recognition'} ($1.91\%$), \textit{`Color'} ($11.82\%$), \textit{`Other Attributes'} ($1.73\%$), \textit{`Activity Recognition'} ($0.52\%$), \textit{`Positional Reasoning'} ($2.32\%$), \textit{`Object Recognition'} ($5.66\%$), \textit{`Absurd'} ($22.16\%$), \textit{`Utility \& Affordance'} ($0.03\%$), \textit{`Object Presence'} ($39.73\%$), \textit{`Counting'} ($9.96\%$) and \textit{`Sentiment Understanding'} ($0.13\%$). Total $1.6$ million question, image and answer triplets are split into \textit{train} and \textit{validation} sets. The \textit{train} set consists of $1.1$ million triplets and $0.5$ million triplets are in the validation split. To deal with language prior issues, TDIUC consists of a special category `Absurd', where an input question is not related to the visual content of a given image. 

% Evaluation Metrics
\subsection{Evaluation Metrics}
\label{subsec:evalMet}

For evaluation of the TDIUC dataset, \textit{Arithmetic-Mean Per Type (AMPT)} and \textit{Harmonic-Mean Per Type (HMPT)} are proposed in \cite{kafle2017analysis} as fair evaluation metrics along with \textit{Overall Accuracy}. The AMPT is the average of question category-wise accuracies with uniform weight to each category. On the other hand, HMPT measures the ability of the model to have a high score across all question types.

The VQA2.0 dataset evaluation is performed using the following metric defined in \cite{antol2015vqa}.

\begin{equation}
     \textbf{Accuracy}{(\mathbf{\hat{a}})} = min\Big\{\frac{\textbf{\#humans that said $\mathbf{\hat{a}}$  }}{\textbf{3}},\textbf{1}\Big\}
\end{equation}

Each question in the VQA2.0 dataset was answered by $10$ annotators. The above evaluation metric considers a predicted answer correct if it matches the answers given by at least $3$ annotators.

% Implementation Details
\subsection{Implementation Details}
\label{subsec:impDet}

Visual feature representation $\mathbf{rI}$ is constructed by extracting $n_v=36$ (for TDIUC) and $n_v=100$ (for VQA2.0) image regions. The use of ResNet-101 embeddings provide image region features of $d_v =2048$ dimensions. The question length in terms of number of tokens ($n_w$) is set to $14$ by trimming or padding. The GloVe word embeddings of $d_w=300$ dimensions are considered. The image and word features are projected to same dimensions $d= 512$. For self- and co-attention computations, the key, query and value vector dimensions are set to $64$, i.e., $d_{KQ} = d_{VS} = 64$. The model uses $n_h = 8$ heads for multi-head attention. The model is trained for $15$ epochs with a batch size of $64$ samples for both experiments and analysis. The hidden layer dimension of answer prediction FCNet is set to $d_{hp} = 1024$. The Adamax optimizer \cite{kingma2014adam} is used with a decaying step learning rate. The initial learning rate is set to $\mathrm{0.002}$, and it decays by $\mathrm{0.1}$ after every $\mathrm{5}$ epochs. The proposed model CSCA is built on the PyTorch framework and is trained on NVIDIA-GTX $\mathrm{1080}$ GPU.

\section{Results and Discussion}
\label{sec:res}

% Comparison with State-of-Art Methods
\subsection{Quantitative Results}
\label{subsec:comp}
\begin{table}[!b]
			\centering
		    \setlength{\tabcolsep}{5pt}
			\caption{Category-wise comparison of CSCA with previous state-of-the-art methods on the TDIUC dataset}
			\begin{tabular}{l|c|c|c|c|c|c}
				\hline
			\textbf{Question Type}	&\textbf{SAN} & \textbf{RAU} & \textbf{MCB} & \textbf{QTA} & \textbf{BAN} & \textbf{CSCA}  \\
			&\textbf{\cite{yang2016stacked}}&\textbf{\cite{kafle2017analysis}} & \textbf{\cite{gao2016compact}} & \textbf{\cite{shi2018question}} & \textbf{\cite{kim2018bilinear}}  &   \\
				\hline
			{Scene Recognition}& 92.3 & 93.96 & 93.06 & 93.80 & 93.1 & \textbf{94.48} \\
			{Sport Recognition}& 95.5 & 93.47 & 92.77 & {95.55} &  95.7 & \textbf{95.85} \\
			{Color Attributes}&60.9 & 66.86 & 68.54 & 60.16 & 67.5 &   \textbf{75.51} \\
			{Other Attributes}&46.2 & 56.49 & 56.72 & 54.36 & 53.2 & \textbf{60.89} \\
			{Activity Recognition}&51.40 & 51.60 & 52.35 & 60.10 & 54.0 &  \textbf{61.00} \\
			{Positional Reasoning}&27.9 & 35.26 & 35.40 & 34.71 & 27.9 & \textbf{42.14} \\
			{Object Recognition}&87.50 & 86.11 & 85.54 & 86.98 & 87.5 &  \textbf{89.11} \\
			{Absurd}&93.4 & 96.08 & 84.82 & \textbf{100.0} &  94.47 & 97.28 \\
			{Utility \& Affordance}& 26.3 & 31.58 & {35.09} & 31.48 & 24.0 & \textbf{40.35} \\
			{Object Presence} & 92.4& 94.38 & 93.64 & 94.55   & 95.1 &\textbf{96.34}\\
			{Counting} &52.1& 48.43 & 51.01 & 53.25 & 53.9 &\textbf{60.70} \\
			{Sentiment Und.} &53.6& 60.09 & 66.25 & 64.38 & 58.7 & \textbf{67.19} \\
				\hline
			{\textbf{Overall Accuracy}}&82.0 & 84.26 & 81.86 & 85.03 & 85.5 & \textbf{88.12} \\
			\textbf{{Harmonic Mean}}&53.7 & 59.00 & 60.47 & 60.08 & 54.9 & \textbf{67.05} \\
			\textbf{Arithmetic Mean} &65.0 & 67.81 & 67.90 & 69.11 & 67.4 & \textbf{73.34} \\
				\hline
			\end{tabular}
			\label{tab:classPerfTDIUC}
\end{table}

\begin{table}[!ht]
\centering
\addtolength{\tabcolsep}{6pt}   
\caption{Comparing \textit{Overall Accuracy} of CSCA for TDIUC dataset}
\begin{tabular}{l|c|c}
\hline
\textbf{Model} & \textbf{Overall Accuracy} & \textbf{Arithmetic Mean}\\
\hline 
\textbf{BTUP\cite{anderson2018bottom}} & 82.91 & 68.82\\
\textbf{QCG\cite{norcliffe2018learning}}  & 82.05 & 65.67\\
\textbf{BAN2-CTI\cite{do2019compact}} & 87.00 & 72.5\\
\textbf{DFAF\cite{gao2019dynamic}} & 85.55 & \textbf{NA}\\
\textbf{RAMEN\cite{shrestha2019answer}}& 86.86 & 72.52\\
\textbf{MLIN\cite{gao2019multi}}& 87.60 & NA \\
\hline
\textbf{CSCA}&\textbf{88.12} & \textbf{73.34}\\
\hline
\end{tabular}
\label{tab:overallPerfTDIUC}
\end{table}	

\textbf{Overall Performance \& Category-wise Performance Comparison on TDIUC Dataset --} Table \ref{tab:classPerfTDIUC} and \ref{tab:overallPerfTDIUC} present the respective class-wise and overall performance for the TDIUC dataset. In terms of the overall accuracy, Arithmetic-MPT (AMPT) and Harmonic-MPT (HMPT) measures, the proposed model CSCA exhibits better performance compared to most of the baseline methods.
Also, in terms of class-wise accuracy, CSCA leads in all except one class. A significant relative gain of 12.6\% is observed compared to the next best performing model for the \textit{`Counting'} category of questions. Table~\ref{tab:woabs} presents the results for different models trained \textit{`Without Absurd'} category of questions. It is observed that CSCA performs better than the existing ones for all three defined evaluation metrics.

\begin{table}[!htbp]
\centering
\caption{Performance of CSCA on TDIUC data (except Absurd category samples) trained without `Absurd' Category samples}
% \addtolength{\tabcolsep}{0.05pt}    
\begin{tabular}{l|c|c|c|c|c}
% \hline
% \multicolumn{6}{c}{\textbf{Without Absurd}}\\
\hline
\textbf{Metrics} & \textbf{MCB} &\textbf{QTA} & \textbf{BAN}& \textbf{BAN2-CTI}&\textbf{CSCA}\\
 & \textbf{\cite{gao2016compact}} &\textbf{\cite{shi2018question}} & \textbf{\cite{kim2018bilinear}}&\textbf{\cite{do2019compact}}& \\
\hline
\textbf{Overall Accuracy} & 78.06 & 80.95& 81.9&85.0&\textbf{85.30}\\
\textbf{Arithmetic-MPT} & 66.07 & 66.88& 64.6 & 70.6 &\textbf{71.21}\\
\textbf{Harmonic-MPT} &55.43 & 58.82 & 52.8 &63.8&\textbf{65.40}\\
\hline
\end{tabular}
\label{tab:woabs}
\end{table}

\textbf{Overall Performance \& Category-wise Performance Comparison on VQA2.0 Dataset --} Table \ref{tab:classPerfVQA2.0} demonstrates the results on test-dev and test-std splits of the VQA2.0 dataset. Performance of the proposed model CSCA is comparable with that of the best among the existing methods. The models LXMERT \cite{tan2019lxmert}, ViLBERT \cite{lu2019vilbert} are pre-trained for multiple vision and language based tasks and are fine-tuned for VQA. Here, CSCA has obtained 67.36\% accuracy on the validation set. This is around 1\% improvement over the best performance among the existing methods.
\begin{table}[!t]
    \centering
		 \setlength{\tabcolsep}{2pt} 
		\caption{Model performance on VQA 2.0 dataset: Validation, Test-Dev \& Test-Std splits. CSCA is compared with several state-of-the-art methods including \textit{Fusion based}, \textit{Visual Attention}, \textit{Dense Attention} based methods separated with lines.}
		\begin{tabular}{l|c|cccccc|cc}
		\hline
		\multirow{2}{*}{\textbf{Methods}}	    &   \textbf{Val}         && \multicolumn{4}{c}{\textbf{Test-Dev}} && {\textbf{Test-Std}} \\
		\cmidrule{2-2}	\cmidrule{4-7} \cmidrule{9-9}    &   \textbf{Overall}         &&     \textbf{Yes / No}        &   \textbf{Number}            &  \textbf{Other}      &   \textbf{Overall}   &&   \textbf{Overall}   &    \\
		\hline
	\textbf{MCB \cite{fukui-etal-2016-multimodal}}       &      59.14          &&      78.46             &   38.28               &    57.80      &   62.27       &&     53.36     \\
	\textbf{MLB \cite{kim2016hadamard}}        &      62.98          &&      83.58             &   44.92               &    56.34      &   66.27       &&      66.62     \\
	\textbf{MUTAN \cite{ben2017mutan}}         &      62.71          &&      82.88             &   44.54               &    56.50      &   66.01       &&       66.38     \\
	\textbf{MFH \cite{yu2018beyond}}        &      62.98          &&      84.27             &   49.56               &    59.89      &   68.76       &&     --    \\
	\textbf{BLOCK \cite{ben2019block}}      &      64.91          &&      83.14             &   51.62               &    58.97      &   68.09       &&    68.41     \\
	\hline
	\textbf{SAN \cite{yang2016stacked}}        &      61.70          &&      78.40             &   40.71               &    54.36      &   61.70       &&   --             \\
    \textbf{BTUP \cite{anderson2018bottom}}    &      63.20          &&      81.82             &   44.21               &    56.05      &   65.32       &&    65.67      \\
    \textbf{BAN \cite{kim2018bilinear}}        &      65.81           &&      82.16             &   45.45               &    55.70      &   64.30       &&      --     \\ 
    \textbf{v-VRANet \cite{yu2020reasoning}}   &      --         &&      {83.31}    &   {45.51}      &   58.41       &   67.20       &&       67.34    \\
    \textbf{ALMA \cite{liu2020adversarial}}        &      --         &&       84.62             &   47.08               &    58.24      &   68.12       &&         66.62     \\
	\textbf{ODA \cite{zhu2021object}}        &  64.23        &&       83.73          &   47.02         &    56.57      &   66.67       &&           66.87            \\
	\textbf{BAN2-CTI \cite{do2019compact}}    &      66.00          &&      --            &   --              &   --     &   --      &&       67.4   \\
    \textbf{CRANet \cite{peng2019cra}}   &         --      &&      {83.31}    &   {45.51}      &   58.41       &   67.20       &&        67.34    \\
    \textbf{CoR \cite{wu2018chain}}        &   65.14         &&      84.98             &   47.19               &    58.64      &   68.19       &&      68.59     \\
	\textbf{MUREL \cite{cadene2019murel}}         &   65.14             &&      84.77             &   49.84               &    57.85      &   68.03       &&        68.41     \\
	\hline
	\textbf{DFAF \cite{gao2019dynamic}}         &   {66.66}             &&      86.09             &   {53.32}              &    {60.49}      &   {70.22}       &&          {70.34}     \\
	\textbf{MLIN \cite{gao2019multi}}         &   66.53             &&      85.96         &    52.93                    &   60.40      &   70.18        &&        70.28     \\
	\textbf{LXMERT \cite{tan2019lxmert}}         &   --             &&      --         &    --                    &   --      &   --       &&        \textbf{72.5}     \\
	\textbf{ViLBERT \cite{lu2019vilbert}}         &   --             &&               &                       &         &   70.55        &&        {70.92}     \\
		\hline
    \textbf{CSCA}            &      \textbf{67.36}         &&      {86.57}          &   \textbf{53.58}              &  \textbf{61.06}   &   \textbf{70.72} &&      {71.04}      \\
		\hline
		\end{tabular}
		\label{tab:classPerfVQA2.0}
\end{table}

\begin{figure*}[!b]%
    \centering
    \subfloat[]{\includegraphics[width=0.4\linewidth]{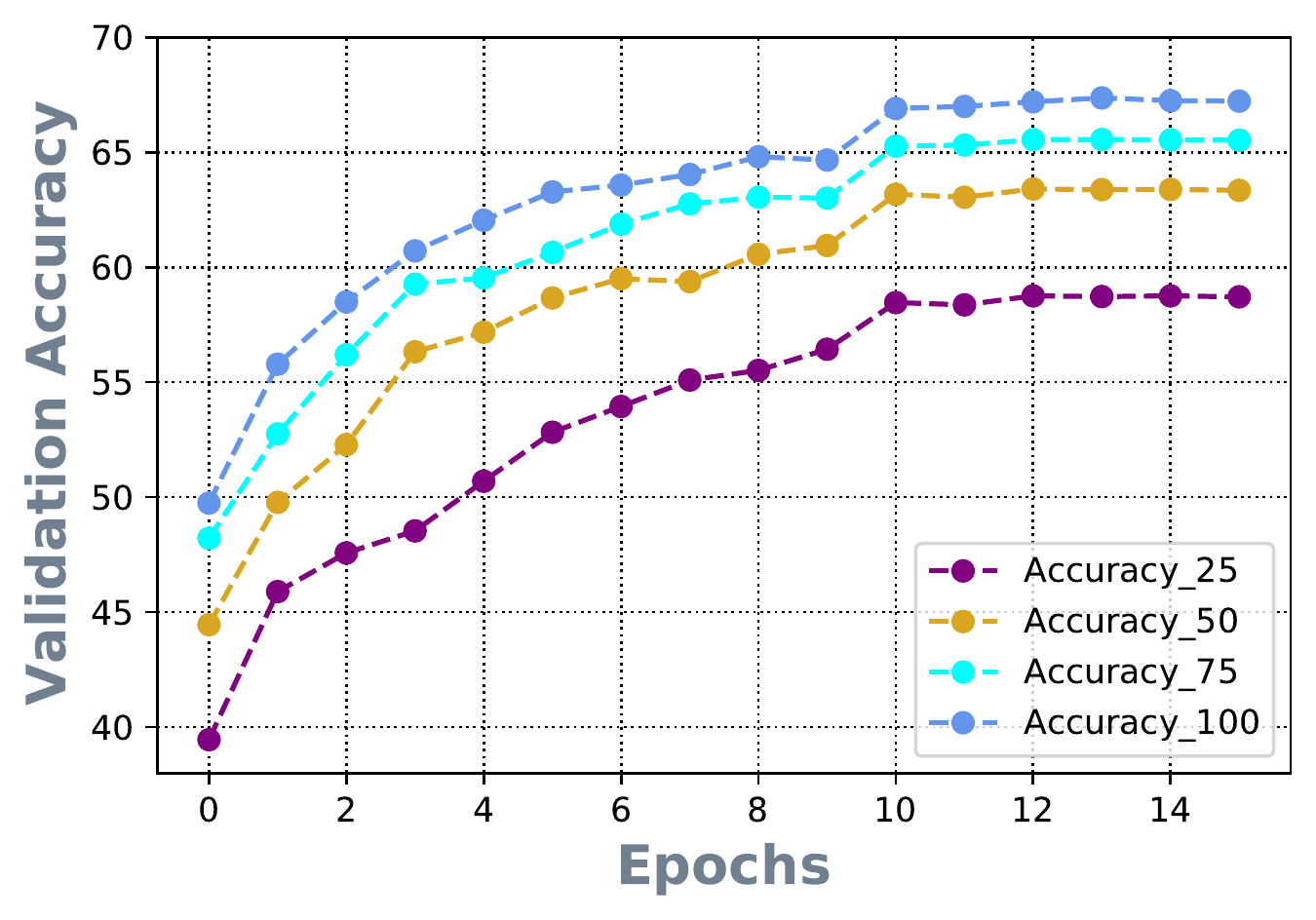}\label{fig:epochAcc}}
    \qquad
    \subfloat[]{\includegraphics[width=0.4\linewidth]{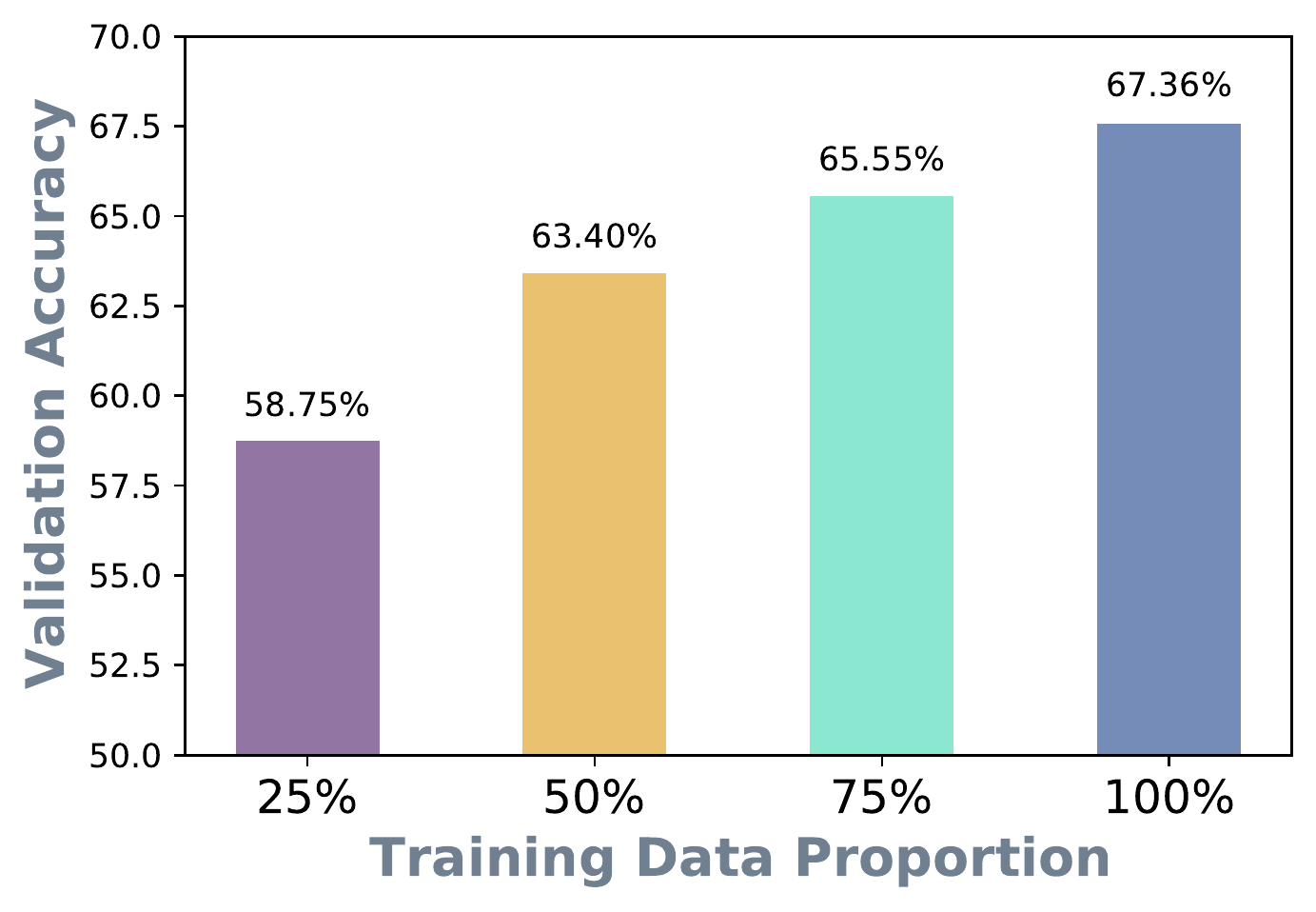}\label{fig:barAcc} }
    \caption{{Illustrating the learning curves on training datasets formed with different amounts of instances from VQA2,0. (a) Performance on validation set of VQA2.0 with respect to the number of epochs. (b) Overall accuracy for VQA2.0 dataset with different proportion of the training data.}}%
    \label{fig:learningCurve}%
\end{figure*}
\subsection{Basic Analysis}
\label{sec:basicAnalysis}
\textbf{Effect of Training Data Size on Performance --}
An analysis is performed to observe the effect of the variation of training dataset size on model performance. The primary objective of this experiment was to ascertain whether a model trained on a smaller dataset can provide similar performance as the one learned from the complete set. To explore this, the model is trained with four different datasets obtained from the original VQA2.0 dataset. The first three datasets are obtained by random shuffling of all samples of the VQA2.0 dataset followed by the extraction of 25\%, 50\% and 75\% samples. The fourth one is the complete VQA2.0 dataset (i.e. 100\%). Other experimental setups like hidden dimension, number of answer classes are kept similar to the original setup for all variants of the dataset. The Epoch-wise performances for the four different datasets are shown in Figure~\ref{fig:epochAcc}. As expected, the model performance improved with an increase in training dataset size. It can be observed that in all four settings, the model performance evolves over a different number of epochs in a similar fashion. However, Figure~\ref{fig:barAcc} indicates that the relative gain achieved by increasing the training dataset size from $25\%$ to $50\%$ is significant compared to that by increasing from $50\%$ to $75\%$ or $75\%$ to $100\%$. This observation may be attributed to the fact that in a collection of randomly shuffled datasets, not many novel instances were encountered during the subsequent increase of the training data.

\begin{figure*}[t]
\centering
\subfloat[]{\includegraphics[width=0.4\linewidth]
{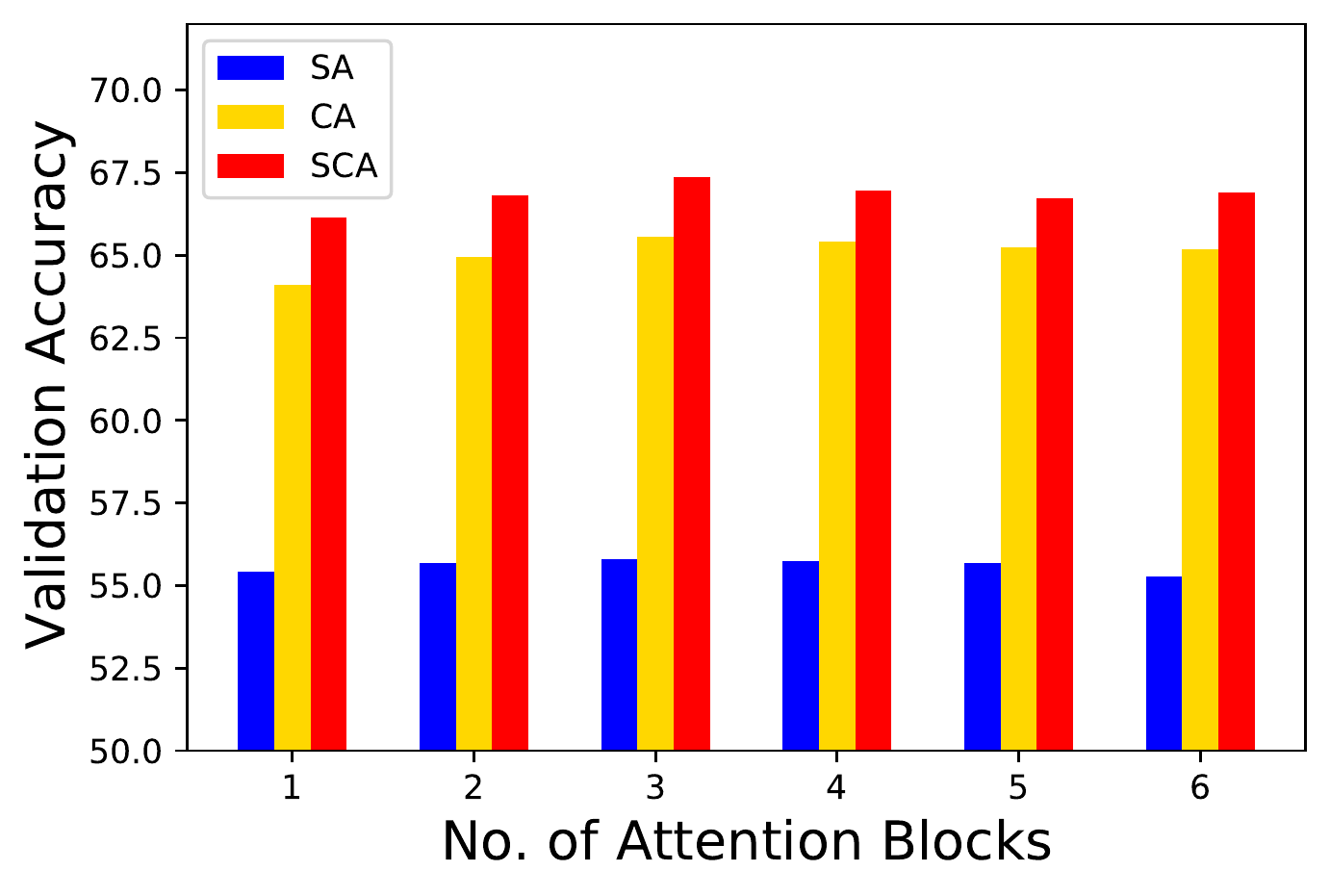}\label{fig:step1}}
\qquad
\subfloat[]{\includegraphics[width=0.4\linewidth]
{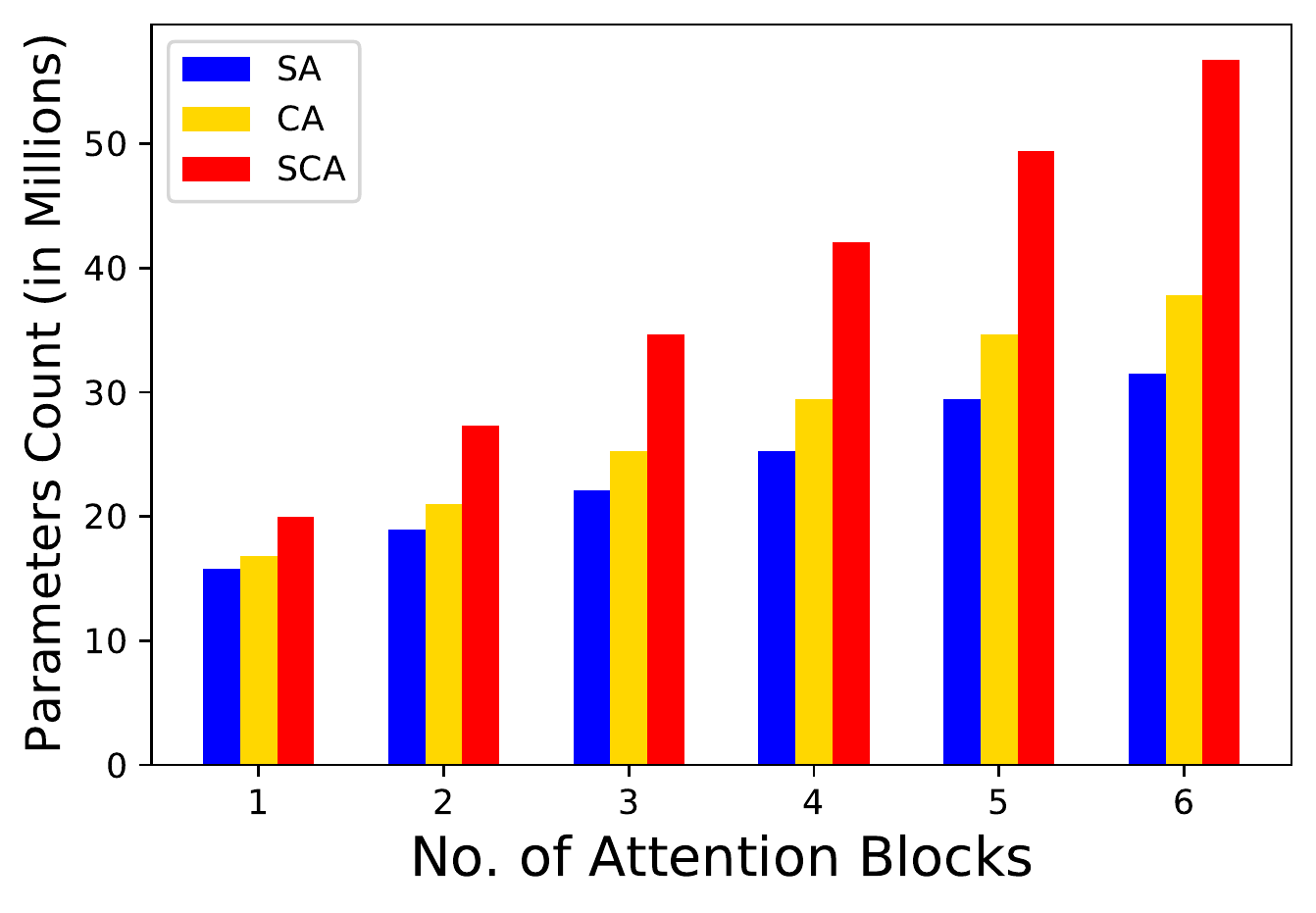}\label{fig:step2}}
\caption{Number of attention blocks incorporated. (a) Validation accuracy for VQA2.0 \textit{`val'} split with respect to attention blocks. (b) Parameter counts with respect to attention blocks}
\end{figure*}

\begin{figure}[H]
\center
{\includegraphics[width=0.4\textwidth,height=8cm,keepaspectratio]{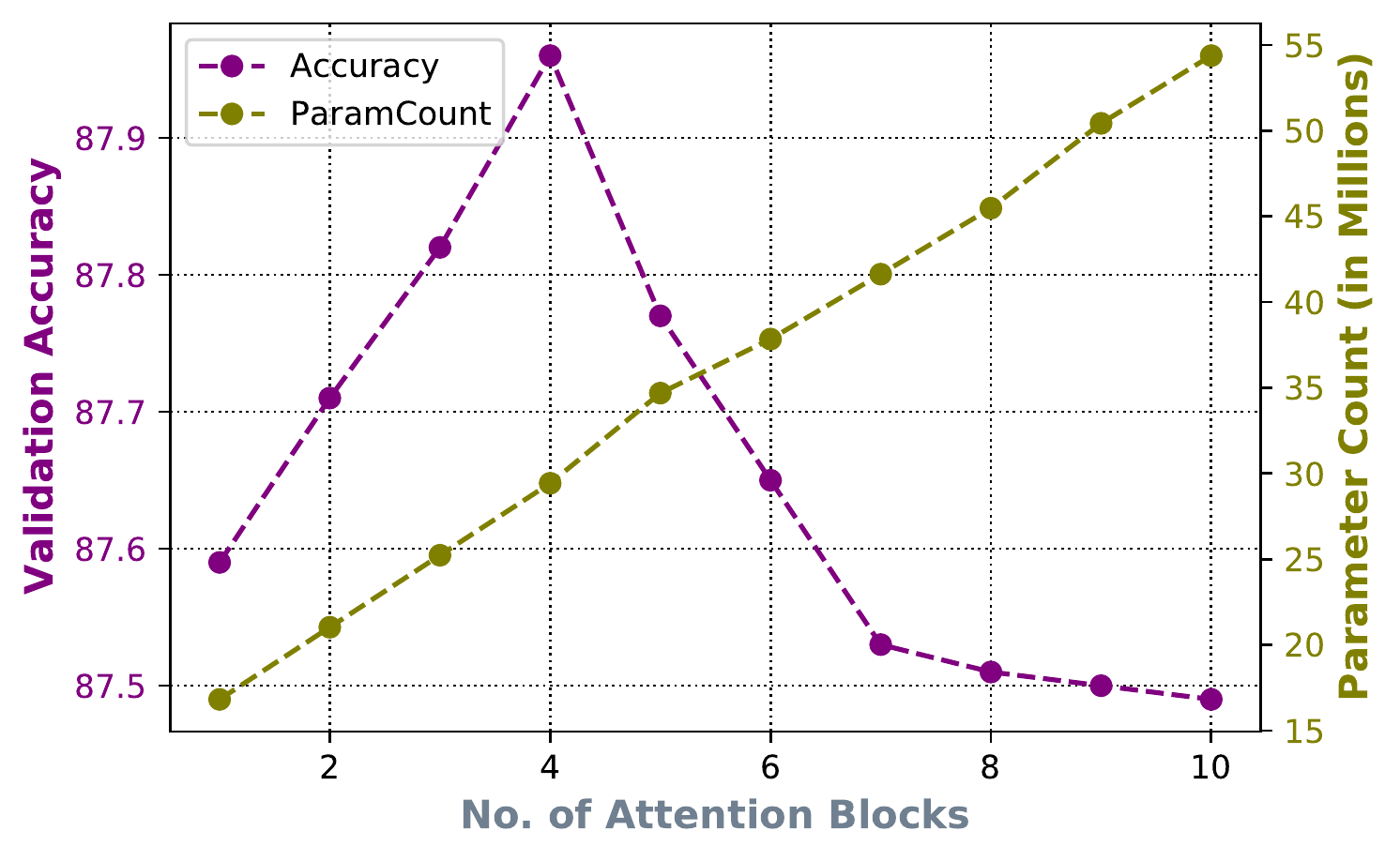}}
\caption{\textit{Validation Accuracy} and {Parameter Count (in Millions)} for \textit{TDIUC} dataset with respect to the number of SCA blocks incorporated in the VQA model.} 
\label{fig:stepATDIUC}
\end{figure}

\textbf{Effect of Number of SCA Blocks -- } In one pass, it is difficult for a model to grasp all relevant information through a representation. Thus, attention blocks in cascade extract the fine-grained information and pass it on to the next one for further refinement. A set of experiments are performed to identify the optimal number of blocks in the cascade. Additionally, the effect of different independent attention mechanisms (SA only, CA only, SCA) for answer prediction is also analyzed. In Figure~\ref{fig:step1}, overall performance for validation split of VQA2.0 dataset is given with respect to varying number of blocks. Figure~\ref{fig:step2} shows the parameter counts with respect to the number of blocks. As per expectation, it is observed that the models perform poorly with single attention blocks (SA only, CA only, SCA). However, the performance is observed to rise only up to four number of blocks. Increasing the number of blocks beyond four does not lead to any further performance improvement. However, adding more blocks also lead to an increase in the number of model parameters (Figure \ref{fig:step2}. Furthermore, one can observe that only CA module can perform better than using only the SA module. This is as per the expectation. Similarly, Figure~\ref{fig:stepATDIUC} shows that the model performance keeps improving until the fourth SCA block for the TDIUC dataset. The model performance starts deteriorating with a further increase in the number of blocks.
% Ablation Studies
\subsection{Ablation Analysis}
\label{subsec:abAnls}
The proposed model performs self-attention on the two modalities to obtain intra-modality correlated features. Then the co-attention module uses respective representations of the two modalities to obtain cross-modality correlated features by performing attention for one modality in the context of another. In this ablation analysis, we examine the impact of individual attention module in various combinations to understand their importance. We also analyze the set of correct predictions obtained in these settings.

\begin{table}[!htbp]
\setlength{\tabcolsep}{5pt}
\centering
\caption{{Evaluating model performance on VQA2.0 dataset to investigate the effect of \emph{different basic attention modules} of the proposed model}}
\begin{tabular}{lc|ccccc}
\hline
\textbf{SA} &   \textbf{CA} &   \textbf{Yes / No} &   \textbf{Number} & \textbf{Other} &  \textbf{Overall}   & \textbf{Parameter}\\
 &    &    &    &  &  \textbf{Accuracy}   & \textbf{(in Millions)}\\
\hline
\ding{55} & \ding{55}        &   69.95   &   36.42   &   50.19    &   55.80  & 22  \\
\ding{51} & \ding{55}       &   79.08   &   40.75   &   49.96    &   59.69 & 15   \\
\ding{55} & \ding{51}       &   81.17   &   44.63   &   56.34    &   64.13   & 25 \\
\ding{51} & \ding{51}       &   \textbf{84.92}   &   \textbf{49.51}   &   \textbf{58.71}    &   \textbf{67.36}  & \textbf{42}  \\
\hline
\end{tabular}\label{tab:ablV} 
\end{table}

\begin{table}[!htbp]
\setlength{\tabcolsep}{10pt}
\centering
\caption{{Evaluating model performance on TDIUC dataset to investigate the effect of \emph{number of attention blocks} and {self-attention \& cross attention.}}}
\begin{tabular}{lc|ccccc}
\hline
\textbf{SA} &   \textbf{CA} &  \textbf{Overall} & \textbf{Parameter}\\
  & & \textbf{Accuracy}   & \textbf{(in Millions)}\\
\hline
\ding{55} & \ding{55}       &    69.18 & 7   \\
\ding{55} & \ding{51}       &    70,46  & 21  \\
\ding{51} & \ding{55}       &    87.42   & 25 \\
\ding{51} & \ding{51}       &   \textbf{88.12}  & \textbf{36}  \\
\hline
\end{tabular}\label{tab:ablT} 
\end{table}

Table~\ref{tab:ablV} and \ref{tab:ablT} present the results of ablation analysis experiments in terms of performance and complexity. The complexity is expressed in terms of the number of model parameters. The first row of the table shows the model performance when neither of the attention is incorporated. The features for both modalities are fused directly via element-wise multiplication without applying self- or co-attention. Second row shows the performance when \emph{only self-attention} (SA only) is incorporated on both modalities and answer prediction is based on the fused embedding of the self-attended representations of the individual modalities. Here, the fused representation is obtained via element-wise multiplication. Third row shows the results when \emph{only co-attention} (CA only) is incorporated on image and question in the context of the other. The last row shows the results from the proposed model that comprises of both \textit{self-attention} and \textit{co-attention} in cascade (SCA). 

As per expectation, the model without any attention mechanism provides the lowest performance (first row). The ``SA only'' model provides lower performance as it lacks the interaction of two modalities and learns a comparatively poor representation (second row). Co-attention is the crucial component for multi-modality that is found to perform better than \textit{self-attention}. In terms of computational complexity, a simple fusion-based model uses the least number of parameters, while the proposed model (SCA) requires the highest number of parameters. However, the performance improvement, especially for VQA2.0 dataset, overcomes the complexity issue. We observe that the change in model performance is similar for both datasets in this analysis.

\begin{figure*}[!t]%
    \centering \subfloat[`Number' Questions]{\includegraphics[width=0.35\linewidth]{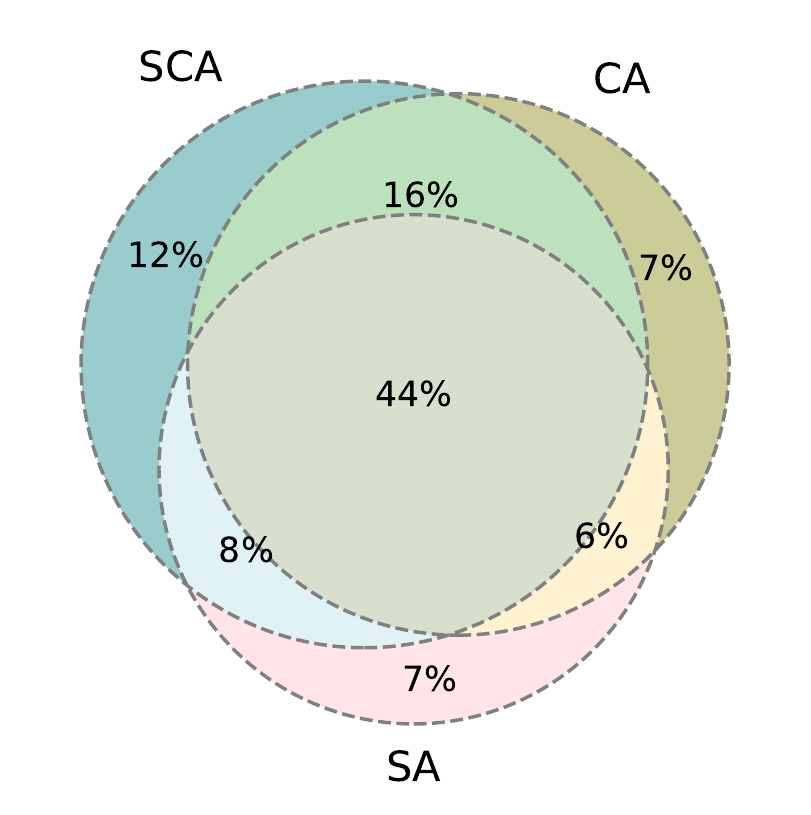}\label{fig:all_num}}
    \subfloat[`Other' Questions]{\includegraphics[width=0.35\linewidth]{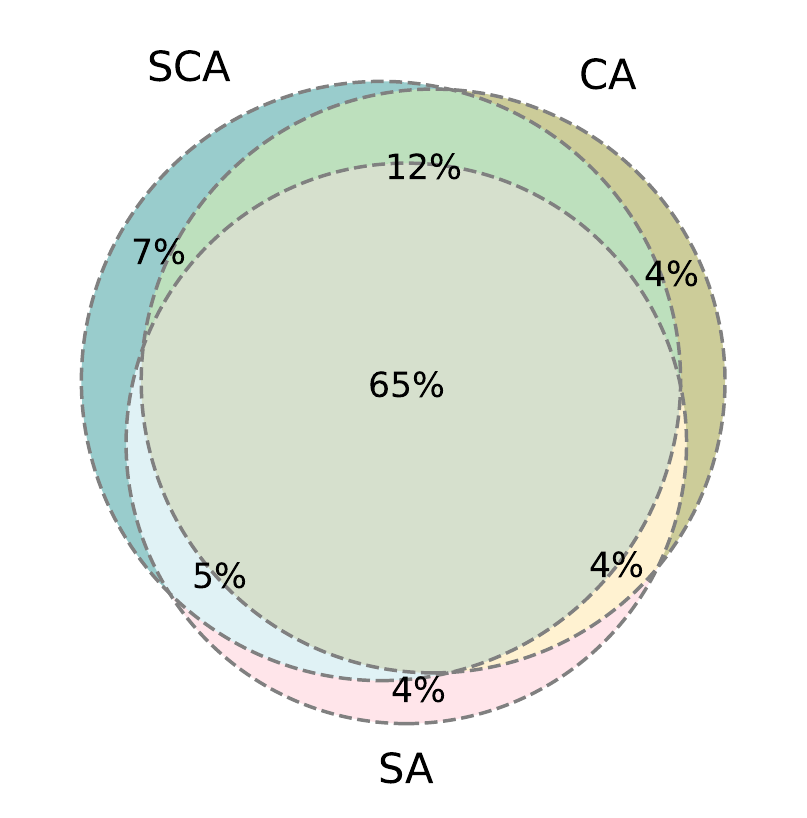}\label{fig:all_oth}}
    \subfloat[`Yes/No' Questions]{\includegraphics[width=0.35\linewidth]{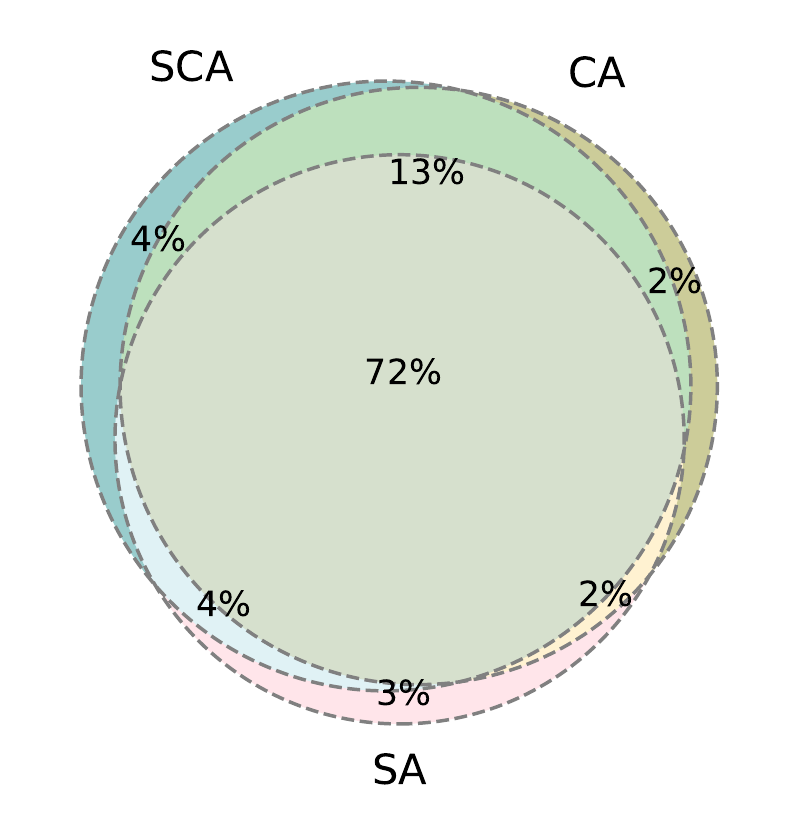}\label{fig:all_bin}}\\
    \caption{{SCA: Self-Attention \& Co-attention, SA: Only self-attention is applied on text and visual features, CA: Cross-Modality Attention on text as well as on visual features guided by each other.}}%
    \label{fig:vennDiag}%
\end{figure*}
% \paragraph{Contributions of various attention mechanism:} 
Figure~\ref{fig:vennDiag} shows the model's performance over various attention mechanisms for the different types of questions category on VQA2.0 dataset. The following are observed from the results for the \emph{`Number'} category of questions. While using the SA only and CA only blocks, the respective models show the overall performances of $65\%$ and $73\%$. Models using SA and CA attention individually predicts $7\%$ of samples correctly that are not correctly classified by any of the other models. Similarly, the model using SCA block classifies $12\%$ of samples correctly that are not correctly classified either by the models using SA or CA only. Thus, the models using SCA blocks achieved the best performance. The same pattern was observed over the other questions types i.e., \emph{`Yes/No'} and \emph{`Other'}. The detailed result for all the question types are shown in figure~\ref{fig:vennDiag}. 
\subsection{Qualitative Results}
\label{subsec:qualRes}
\begin{figure*}
\centering
\begin{subfigure}[b]{0.2\textwidth}
% \centering
\includegraphics[width=3cm, height = 3cm]{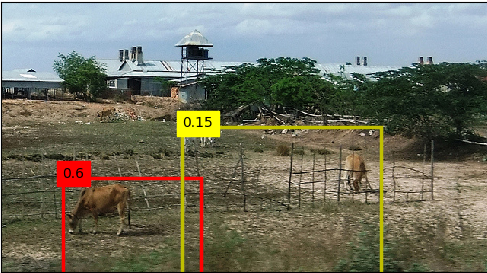}
\caption{\textbf{Q.} Are there \textcolor{blue}{any} cows in the picture ?\\ \textbf{ Ans: } Yes \textcolor{violet}{\ding{52}} (\textcolor{red}{0.59}, \textcolor{orange}{0.21}) \\ \textbf{GT: }Yes}\label{fig:qual1}
\end{subfigure}%
\hspace{0.2in}
\begin{subfigure}[b]{0.2\textwidth}
% \centering
\includegraphics[width=3cm,height= 3 cm]{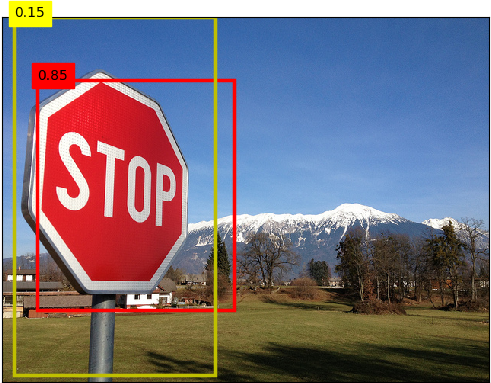}
\caption{\textbf{Q.} What shape is the \textcolor{blue}{stop} sign ? \\ \\ \textbf{Ans:}Octagon \textcolor{violet}{\ding{52}}(\textcolor{red}{0.85}, \textcolor{orange}{0.15}) \\ \textbf{GT: }Octagon}\label{fig:qual2}
\end{subfigure}%
\vspace{0.1in}
\begin{subfigure}[b]{0.2\textwidth}
% \centering
\includegraphics[width=3cm,height= 3 cm]{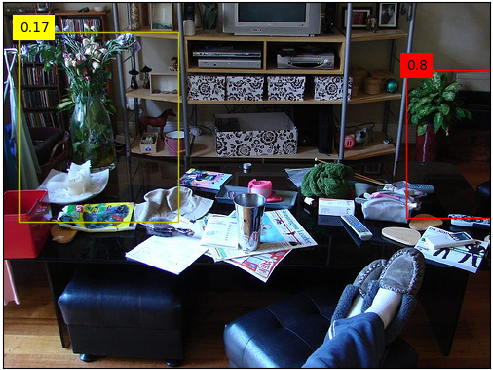}
\caption{\textbf{Q.} How {many} potted \textcolor{blue}{plants} ? \\ \\ \textbf{ Ans: } 2 \textcolor{violet}{\ding{52}} (\textcolor{red}{0.80}, \textcolor{orange}{0.17}) \\ \textbf{GT: }2}\label{fig:qual3}
\end{subfigure}%
\hspace{0.2in}
\begin{subfigure}[b]{0.2\textwidth}
% \centering
\includegraphics[width=3cm,height= 3 cm]{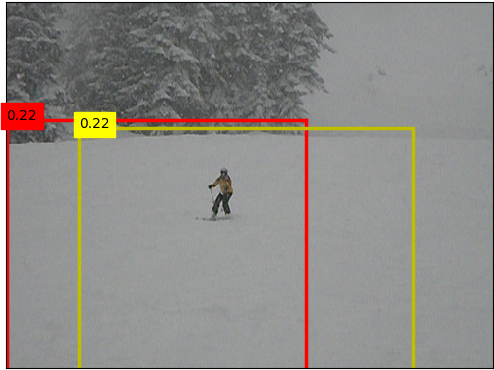}
\caption{\textbf{Q.} What is the color of the \textcolor{blue}{ground} ? \\ \textbf{ Ans: } White \textcolor{violet}{\ding{52}} (\textcolor{red}{0.22}, \textcolor{orange}{0.22}) \\ \textbf{GT: }White}\label{fig:qual4}
\end{subfigure}%
\vspace{0.1in}
\begin{subfigure}[b]{0.2\textwidth}
% \centering
\includegraphics[width=3cm,height= 3 cm]{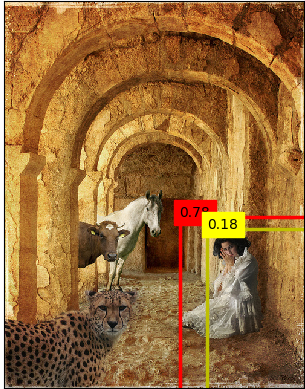}
\caption{\textbf{Q.} What is the lady \textcolor{blue}{doing} ? \\ \textbf{ Ans: } Sitting \textcolor{violet}{\ding{52}} (\textcolor{red}{0.78}, \textcolor{orange}{0.18}) \\ \textbf{GT: }Sitting}\label{fig:qual5}
\end{subfigure}%
\hspace{0.2in}
\begin{subfigure}[b]{0.2\textwidth}
% \centering
\includegraphics[width=3cm,height= 3 cm]{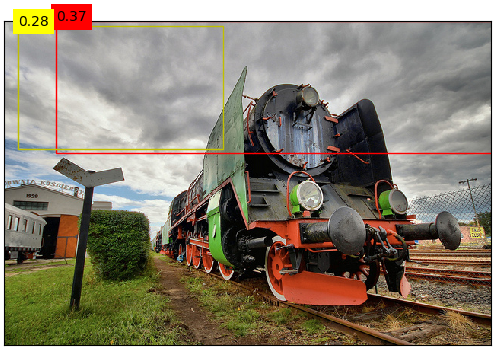}
\caption{\textbf{Q.} What is the weather \textcolor{blue}{like} ? \\ \textbf{ Ans: } Cloudy \textcolor{violet}{\ding{52}} (\textcolor{red}{0.37}, \textcolor{orange}{0.28}) \\ \textbf{GT: }Cloudy}\label{fig:qual6}
\end{subfigure}%
\vspace{0.1in}
\begin{subfigure}[b]{0.2\textwidth}
% \centering
\includegraphics[width=3cm,height= 3 cm]{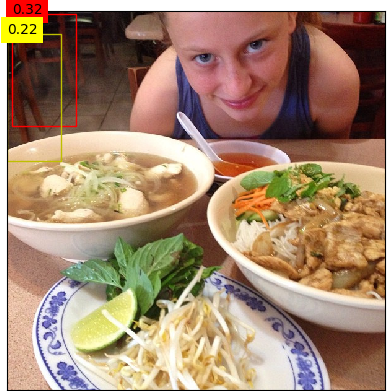}
\caption{\textbf{Q.} What \textcolor{blue}{colors} are the chair ? \\ \textbf{ Ans: } Brown \textcolor{violet}{\ding{52}} (\textcolor{red}{0.78}, \textcolor{orange}{0.18}) \\ \textbf{GT: }Brown}\label{fig:qual7}
\end{subfigure}%
\hspace{0.2in}
\begin{subfigure}[b]{0.2\textwidth}
% \centering
\includegraphics[width=3cm,height= 3 cm]{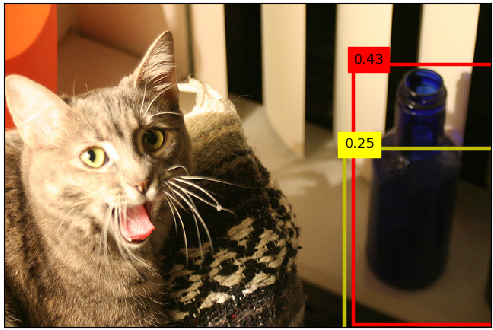}
\caption{\textbf{Q.} Are there any \textcolor{blue}{bottles} in photo ? \\ \textbf{ Ans: } Yes \textcolor{violet}{\ding{52}} (\textcolor{red}{0.43}, \textcolor{orange}{0.25}) \\ \textbf{GT: }Yes}\label{fig:qual8}
\end{subfigure}%
\caption{Qualitative results for our proposed method CSCA. Attention for image obtained with cascade of $T=4$ SCA blocks is presented. (\textcolor{red}{top1}, \textcolor{orange}{top2}) attention score values correspond to the top two attention weight obtained for top-2 salient regions, that are relevant to infer the answer. The question words shown in blue are the ones that get the highest attention score.}\label{fig:qualRes}
\end{figure*}

The qualitative results are presented in Figure~\ref{fig:qualRes} to demonstrate the efficacy of the proposed model. For this, two salient regions of a given image with the highest attention scores are highlighted. These are the attention scores obtained after cascading $T=4$ SCA blocks. The question words that obtain the highest attention scores are also highlighted. As evident from Figure~\ref{fig:qual1}, the proposed model CSCA is able to focus on relevant image regions and question words. The top-2 salient regions corresponding to the binary question \textit{``Are there any cows in the picture?"} are the ones that capture the \emph{cows} and hence, the model responds by the answer \textit{`Yes'}. Similarly, Figures~\ref{fig:qual2}~\ref{fig:qual3}~\ref{fig:qual4}~\ref{fig:qual5}~\ref{fig:qual6}~\ref{fig:qual7}~\ref{fig:qual8} show that the model is trying to identify the salient image regions and relevant question words to predict the appropriate answer.

 \begin{figure}[!htbp]
\centering
\begin{subfigure}[b]{0.2\textwidth}
\centering
\includegraphics[width=3.25cm,height= 3 cm]{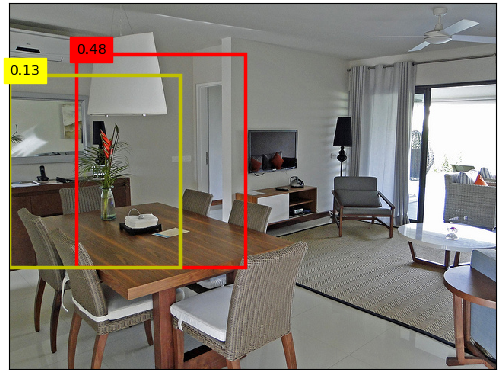}
\caption{\textbf{Q.} What room is shown in the picture ?\\ \textbf{ Ans: } Kitchen \textcolor{red}{\ding{55}} (\textcolor{olive}{0.48}, \textcolor{teal}{0.13}) \\ \textbf{GT: }Living Room}\label{fig:ENE1}
\end{subfigure}%
\hspace{0.2in}
\begin{subfigure}[b]{0.2\textwidth}
\centering
\includegraphics[width=3.25cm,height= 3 cm]{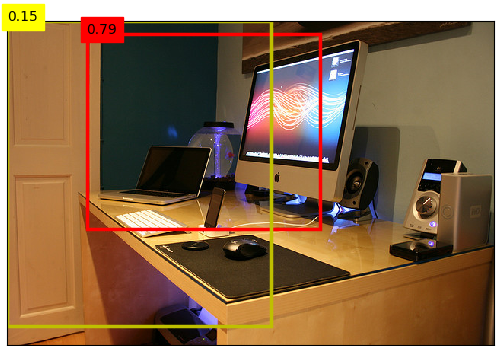}
\caption{\textbf{Q.} What color is the wall in back of the desk ? \\ \textbf{ Ans: } Green \textcolor{red}{\ding{55}} (\textcolor{olive}{0.79}, \textcolor{teal}{0.15}) \\ \textbf{GT: }Gray}\label{fig:ENE2}
\end{subfigure}%
\caption{Failure cases where wrong attention leads to incorrect answer prediction.}
\end{figure}
However, the model made errors as well. One of the reasons was incorrect attention to image regions.
As shown in Figure~\ref{fig:ENE1}, the model's focus is primarily on the position from where it seems like this room is a kitchen. If the attention is given to other regions, the answer will likely to change to  `\textit{living room}'. In~\ref{fig:ENE2} for question `\textit{What color is the wall in back of the desk ?}', the model focuses on the other side of the desk instead of the back. The predicted answer is `green', the color on the side-wall of the desk.

\section{Conclusion}
\label{sec:con}
This work proposes a dense attention mechanism-based VQA model. Dense attention is incorporated by exploiting both self-attention and co-attention. The self-attention mechanism helps in obtaining improved representation within a single modality. With self-attention, a salient region (in the case of image) interacts with every other region. The final representation inherits the contextual information for all regions. Similarly, for the input questions, self-attention provides the representation of every single word that captures the contextual information for other words as well. The proposed model also exploits the cross-modal interaction of two modalities which is further strengthened by self-attention of two modalities. Attention blocks are cascaded multiple times to facilitate refined cues of visual and textual features. The model's capability is justified by detailed experiments and analysis performed on the two benchmark VQA datasets.

%Revised Future Work
The proposed method can be extended in several ways. The present proposal may be subjected to bias and consistency analysis. For example, this may be performed by rephrasing questions and flipping (or rotating) associated images. Also, the current proposal can be extended with translated question-answer pairs to validate its applicability in multi-lingual VQA.

\bibliography{references.bib} 

\newpage

\end{document}